%% file: iclr2025_conference.tex
\documentclass{article} % For LaTeX2e
\usepackage{iclr2025_conference,times}

\input{math_commands.tex}

\usepackage{hyperref}
\usepackage{url}

\usepackage[utf8]{inputenc} % allow utf-8 input
\usepackage[T1]{fontenc}    % use 8-bit T1 fonts
\usepackage{booktabs}       % professional-quality tables
\usepackage{amsfonts}       % blackboard math symbols
\usepackage{nicefrac}       % compact symbols for 1/2, etc.
\usepackage{microtype}      % microtypography
\usepackage{xcolor}         % colors
\usepackage{algorithm}
\usepackage{algpseudocode}
\usepackage{adjustbox}
\usepackage{lipsum}
\usepackage{array}
\usepackage{pgf-pie}
\usepackage{graphicx}
\usepackage{wrapfig}

\usepackage{subcaption}
\usepackage{multirow}
\usepackage{caption}
\usepackage{makecell}
\input{def}

\newtheorem{dfn}{Definition}

\title{MemBench: Memorized Image Trigger Prompt Dataset for Diffusion Models}

\author{%
  Chunsan Hong \\
  KAIST\\
  South Korea\\
  \texttt{hoarer@kaist.ac.kr} \\
  \And
  Tae-Hyun Oh\textsuperscript{$\dagger$}\\
  Dept. EE \& GSAI, POSTECH, \\
  I-CREATE, Yonsei Univ.\\
  South Korea \\
  \texttt{taehyun@postech.ac.kr} \\
  \And
  Minhyuk Sung\textsuperscript{$\dagger$}\\
  KAIST \\
  South Korea \\
  \texttt{mhsung@kaist.ac.kr} \\
}

\iclrfinalcopy % Uncomment for camera-ready version, but NOT for submission.
\begin{document}

\maketitle
\def\thefootnote{$\dagger$}\footnotetext{Co-corresponding authors.}
\input{0_abstract}

\def\thefootnote{\arabic{footnote}}
\setcounter{footnote}{0}

\input{1_introduction}

\input{2_related_work}

\input{3_method}

\input{4_result}

\input{5_membench}

\input{6_findings}

\input{7_evaluation}

\input{8_conclusion}

\section*{Acknowledgments}
T.-H. Oh was partially supported by Institute of Information \& communications Technology Planning \& Evaluation (IITP) grant funded by the Korea government (MSIT) (No.RS-2021-II212068, Artificial Intelligence Innovation Hub).

\section*{Ethics statement}
Our work introduces a technique for extracting the training data of diffusion models. 
This could potentially harm the rights of model owners or image copyright holders. 
Therefore, it is crucial to handle this technique with caution to avoid any infringement issues.
For more details, please refer to Appendix.

\section*{Reproducibility statement}
We provide the code for our training data extraction algorithm, the dataset, and the evaluation in the supplementary material.

\bibliography{iclr2025_conference}
\bibliographystyle{iclr2025_conference}

\end{document}

%% file: math_commands.tex
\usepackage{amsmath,amsfonts,bm}

\def\eqref#1{equation~\ref{#1}}
\def\1{\bm{1}}

\def\rvf{{\mathbf{f}}}

\def\rvp{{\mathbf{p}}}

\def\rvw{{\mathbf{w}}}
\def\rvx{{\mathbf{x}}}

\def\rmI{{\mathbf{I}}}

\def\rmT{{\mathbf{T}}}

\def\vtheta{{\bm{\theta}}}
\def\vphi{{\bm{\phi}}}

\def\veps{{\bm{\epsilon}}}

\DeclareMathAlphabet{\mathsfit}{\encodingdefault}{\sfdefault}{m}{sl}
\SetMathAlphabet{\mathsfit}{bold}{\encodingdefault}{\sfdefault}{bx}{n}

\DeclareMathOperator*{\argmax}{arg\,max}

%% file: def.tex
\newcommand{\matrixb}{\left[ \begin{array}}
\newcommand{\matrixe}{\end{array} \right]}

\usepackage{xspace}
\makeatletter
\DeclareRobustCommand\onedot{\futurelet\@let@token\@onedot}
\def\@onedot{\ifx\@let@token.\else.\null\fi\xspace}

\def\ie{\emph{i.e}\onedot}

\makeatother

\newcommand{\Cref}[1]{Chap.~\ref{#1}}

\renewcommand{\paragraph}[1]{\vspace{1mm}\noindent\textbf{#1}\,\,\,}

\usepackage{enumitem}
\setlist[itemize]{align=parleft,left=0pt}

%% file: 0_abstract.tex
\begin{abstract}
Diffusion models have achieved remarkable success in Text-to-Image generation tasks, leading to the development of many commercial models. However, recent studies have reported that diffusion models often generate replicated images in train data when triggered by specific prompts, potentially raising social issues ranging from copyright to privacy concerns. 
To sidestep the memorization, there have been recent studies for developing memorization mitigation methods for diffusion models. 
Nevertheless, the lack of benchmarks impedes the assessment of the true effectiveness of these methods. 
In this work, we present MemBench, the first benchmark for evaluating image memorization mitigation methods. Our benchmark includes a large number of memorized image trigger prompts in various Text-to-Image diffusion models. Furthermore, in contrast to the prior work evaluating mitigation performance only on trigger prompts, we present metrics evaluating on both trigger prompts and general prompts, so that we can see whether mitigation methods address the memorization issue while maintaining performance for general prompts. This is an important development considering the practical applications which previous works have overlooked. 
Through evaluation on MemBench, we verify that the performance of existing image memorization mitigation methods is still insufficient for application to diffusion models. The code and datasets are available at \href{https://github.com/chunsanHong/MemBench\_code}{https://github.com/chunsanHong/MemBench\_code}
\end{abstract}

%% file: 1_introduction.tex
\section{Introduction}\label{introduction}

Text-to-Image (T2I) generation has shown significant advancements and successes with the advance of diffusion models, such as Stable Diffusion~\citep{rombach2022stablediffusion}.
Compared to previous generative models, diffusion models excel in generating diverse and high quality images from user-desired text prompts, which has led to the vast release of commercial models such as MidJourney. However, recent studies~\citep{somepalli2023diffusion,somepalli2023understanding,carlini2023extracting} have revealed that certain text prompts tend to keep replicating 
images in the train dataset
% the train data images
which can cause serious privacy issues. This issue has already triggered controversy
% been a topic of discussion 
in the real world: specific prompts containing the term ``Afghan'' have been known to reproduce copyrighted images of the Afghan girl when using MidJourney~\citep{wen2024detecting}. One of the major issues with such prompts is that, regardless of initial random noise leveraged in the reverse process of the diffusion model, they always invoke almost or exactly same memorized images~\citep{wen2024detecting,carlini2023extracting,webster2023reproducible}.

% Consequently, 
To address this matter,
\citet{wen2024detecting} and \citet{somepalli2023understanding} have proposed 
% provided
mitigation methods to prevent the regeneration of identical images in the train dataset
% train data images
invoked from certain text prompts.
% in T2I diffusion models.
% A significant challenge, however, lies in the absence of benchmarks to measure the efficacy of these memorization mitigation methods. 
However, the evaluation of these memorization mitigation methods has  lacked rigor and comprehensiveness
due to the absence of benchmarks.
As an adhoc assessment method, 
% The alternative assessment method that
the current studies~\citep{wen2024detecting,somepalli2023understanding} have adopted the following method:
% is to  
1) simulating memorization by fine-tuning T2I diffusion models for overfitting on a separate small and specific dataset of \{image, prompt\} pairs,
% to encourage overfitting,
and 2) assess whether the images used in the fine-tuning are reproduced from the query prompts. However, it remains unclear whether 
% memorization mitigation methods whose efficacy is verified in such fine-tuned models will effectively resolve the memorization issues in existing diffusion models.
such evaluation results can be extended to practical scenarios with the pre-trained existing diffusion models and represent the effectiveness for resolving memorization.

In this work, we present \textbf{MemBench}, the first benchmark for evaluating image memorization mitigation methods for diffusion models. Our MemBench includes the following key features to ensure effective evaluation: (1) MemBench provides a sufficient number of memorized image trigger prompts, with 3000 and 1500 prompts for Stable Diffusion 1 and 2, most popular open-source models, respectively. Previous work~\citep{webster2023reproducible} has offered the highest number of prompts at 345 and 218 for Stable Diffusion 1 and 2, respectively; however, we increase the number of prompts to enhance the reliability of the evaluation. We also provide memorized images and trigger prompts for DeepFloydIF~\citep{shonenkov2023deepif} and Realistic Vision~\citep{CivitAI}. (2) We take into account a general prompt scenario, which has been overlooked in prior work. The prior mitigation methods 
% Previous studies~
\citep{wen2024detecting,ren2024unveiling,somepalli2023understanding} 
% offering mitigation methods 
have been evaluated 
% the performance of their methods
solely on memorized image trigger prompts. However, to apply 
% for
mitigation methods in practice,
% to be applicable in real-world scenarios, 
they should also maintain performance on general prompts. (3) We suggest rigorous metrics. As previous mitigation works~\citep{wen2024detecting,somepalli2023understanding} have measured, MemBench also includes the SSCD~\citep{pizzi2022self}, which measures the similarity between memorized and generated images, and the CLIP Score~\citep{hessel2021clipscore}, which measures Text-Image alignment. Additionally, MemBench involves Aesthetic Score~\citep{schuhmann2022laion} to assess image quality, which has been overlooked by prior work and allows to penalize unuseful trivial solutions. 
(4) We propose the 
% ideal
reference performance that mitigation methods should achieve. 
In previous works~\citep{ren2024unveiling,wen2024detecting,somepalli2023understanding}, the effectiveness of mitigation methods has been demonstrated by measuring the decrease in SSCD and the extent to which the CLIP Score is maintained before and after applying the mitigation method. However, this does not necessarily confirm whether image memorization has been adequately mitigated.
Therefore, for SSCD and CLIP Score, we provide guidelines on the target value that should be achieved for a mitigation method to be considered effective.

MemBench reveals several key findings:
\begin{enumerate}
\vspace{-8pt}
\setlength{\itemsep}{-0.2mm}
\setlength{\leftskip}{-8.3mm}
\item All image memorization mitigation methods result in a reduction of Text-Image alignment between generated images and prompts when applied to the generation process of diffusion models.
\item The mitigation methods affect the image generation capabilities of diffusion models, which can lead to lower image quality and the production of less natural images.
\item The mitigation methods can cause performance degradation in the general prompt scenario, which may pose challenges for practical application.
\end{enumerate}

Our additional contribution lies in offering an effective algorithm to search for memorized image trigger prompts. The absence of such benchmarks originates from the significant challenge of collecting prompts that induce memorized images. Compared to the existing methods~\citep{carlini2023extracting,webster2023duplication} that require training data itself~\citep{schuhmann2022laion} for fetching candidate prompts, our proposed based construction method is developed with 
Markov Chain Monte Carlo (MCMC)~\citep{metropolis1953equation}, which enables
the efficient search for these problematic prompts in an open token space without any dataset.

%% file: 2_related_work.tex
\section{Related Work}\label{related_work}
\textbf{Memorization Mitigation Methods.} 
Memorization mitigation methods are divided into two approaches: one is the inference time method, and the other is the train time method. Inference time methods aim to prevent the generation of memorized images that already exist in pretrained diffusion models during the image generation process. \citet{somepalli2023understanding} propose a rule-based text embedding augmentation to mitigate memorization in diffusion models. This includes adding Gaussian noise to text embeddings or inserting random tokens in the prompt. \citet{wen2024detecting} propose the loss that predicts if a prompt will induce a memorized image, and present a mitigation strategy that applies adversarial attacks on this loss to modify the text embeddings of trigger prompts. Both of these works evaluate their methods by intentionally overfitting the diffusion model on specific \{image, text\} pairs to induce the memorization effect, and then checking whether the images are regenerated from the corresponding prompts when their methods are applied. \citet{ren2024unveiling} analyze the impact of trigger prompts on the cross-attention layer of diffusion models and propose a corresponding mitigation method.

Train time methods aim to prevent diffusion models from memorizing training data during model training by employing specific training techniques. Although several methods~\citep{daras2024ambient,liu2024iterativeensembletrainingantigradient} have been proposed, experiments have been conducted only on small models and datasets such as CIFAR-10 and CelebHQ. While some experiments~\citep{somepalli2023understanding,ren2024unveiling,wen2024detecting} have been conducted on large models such as Stable Diffusion, they only assess whether the fine-tuning dataset is memorized when fine-tuning the model. To date, no train time mitigation method has been tested by training large diffusion models from scratch to evaluate its effectiveness.

Therefore, we consider inference time methods as the primary approaches applicable to foundation models like Stable Diffusion. However, due to the absence of a proper benchmark that contains sufficient number of memorized image trigger prompts, evaluation remains inadequate. MemBench addresses this gap by providing sufficient test data and appropriate metrics for the comprehensive evaluation of these inference time methods.

\textbf{Training Data Extraction Attack.}
\citet{carlini2023extracting} propose a method to search for memorized image trigger prompts involving two stages. In the pre-processing stage, they embed the entire training set of Stable Diffusion into the CLIP~\citep{radford2021clip} feature space and cluster these embeddings to identify the most repeated images. In the post-processing stage, Stable Diffusion is used to generate 500 images for each prompt corresponding to these clustered images.  The similarity among these 500 generated images is measured, and only those prompts that produce highly similar images are sampled. Finally, image retrieval is performed on the training data using generated images from these selected prompts to verify if the generated images match the training data images. The pre-processing involves CLIP embedding and clustering of 160M images, while the post-processing involves generating 175M images. \citet{webster2023reproducible} propose an advanced searching algorithm. In the pre-processing stage, an encoder is trained to compress CLIP embeddings. Then, 2B CLIP embeddings are compressed and clustered using KNN~\citep{webster2023duplication}. In the post-processing stage, Webster
% \citet{webster2023reproducible}
introduces an effective method that performs a few inferences of the diffusion model to predict whether a prompt will induce memorized images. This method is applied to 20M prompts acquired from the pre-processing stage. 

Both methods share common bottlenecks: they are memory inefficient and require extremely high computational costs. Additionally, the most fundamental problem is their reliance on training data as candidate trigger prompts. With LAION becoming inaccessible\footnote{\url{https://laion.ai/notes/laion-maintenance/}}, these methods can no longer be reproducible and utilized. However, our method can search more for trigger prompts efficiently than those methods even without any pre-processing steps and any dataset.

\textbf{Benchmark Dataset.}
Since the only existing dataset that can be used for evaluating mitigation methods is the small dataset released by \citet{webster2023reproducible}, 
\citet{ren2024unveiling} evaluate their method on the Webster dataset, while the Webster dataset is not originally purposed as a benchmark dataset.
The dataset is constructed by the training data extraction attack method proposed by Webster, which is not scalable; thus, the dataset remains a small scale.
Also, Ren~et al.~did not measure the loss of semantic preservation after mitigation, which is an important criterion but overlooked.
Our benchmark is also constructed by a novel and efficient data extraction method, and the first benchmark for evaluating those mitigation methods with carefully designed metrics.

%% file: 3_method.tex
\section{Searching Memorized Image Trigger Prompt Leveraging MCMC}\label{method}
\subsection{Preliminary}\label{preliminary}
\paragraph{Diffusion Models.} Denoising Diffusion Probabilistic Model (DDPM)~\citep{ho2020DDPM} is a representative diffusion model designed to approximate the real data distribution \( q(\rvx) \) with a model \( p_\vtheta(\rvx) \). For each \( \rvx_0 \sim q(\rvx) \), DDPM constructs a discrete Markov chain \( \{\rvx_0, \rvx_1, \ldots, \rvx_T\} \) that satisfies \( q(\rvx_t|\rvx_{t-1}) = \mathcal{N}(\rvx_t; \sqrt{1-\beta_t}\rvx_{t-1}, \beta_t\rmI) \). This is referred to as the forward process, where \( \{\beta_t\}_{t=1}^T \) is a sequence of positive noise scales. Conversely, the reverse process generates images according to \( p_\vtheta(\rvx_{t-1}|\rvx_t) = \mathcal{N}(\rvx_t; \mu_\vtheta(\rvx_t, t), \Sigma_\vtheta(\rvx_t, t)) \).
DDPM starts by sampling \( \rvx_T \) from a Gaussian distribution, and then undergoes a stochastic reverse process to generate the sample \( \rvx_0 \), \ie an image. 
With a parametrized denoising network \( \veps_{\vtheta} \), this generation process can be expressed as:
\begin{align}
    \rvx_{t-1} = \tfrac{1}{\sqrt{\alpha_t}}\left(\rvx_t - \tfrac{1-\alpha_t}{\sqrt{1-\bar{\alpha}_t}}\veps_\vtheta(\rvx_t,t)\right)+\sigma_t\rvw,
\end{align}
where $\alpha_t = 1-\beta_t$, $\bar{\alpha}_t=\prod_{i=1}^t\alpha_t$, $\sigma_t$ can be $\sqrt{\beta}$ or $\sqrt{\frac{1-\bar{\alpha_{t-1}}}{1-\bar{\alpha}_t}\beta_t}$, and $\rvw\sim\mathcal{N}(0;\rmI)$. 
The equations may vary depending on hyper-parameter choices and the numerical solver used~\citep{song2022denoising,song2021scoresde}. 

\paragraph{Classifier Free Guidance (CFG).}
In T2I diffusion models such as Stable Diffusion~\citep{rombach2022stablediffusion}, CFG~\citep{ho2022classifier} is commonly employed to generate images better aligned with 
% from 
the desired prompt. Given a text prompt $\rvp$ and the text encoder $\rvf(\cdot)$ of the pre-trained CLIP~\citep{radford2021clip}, predicted noise is replaced as follows:
\begin{align}
\Tilde{\veps}_{\vtheta,\rvf}(\rvx, \rvp, t) = \veps_\vtheta(\rvx,\rvf(\emptyset), t)+s\cdot(\veps_\vtheta(\rvx,\rvf(\rvp),t)-\veps_\vtheta(\rvx,\rvf(\emptyset),t)),
\end{align}
where $\emptyset$ denotes the empty string, and $s$ is the guidance scale. 

\paragraph{Memorized Image Trigger Prompt Prediction.} \citet{wen2024detecting}  proposed an efficient method to predict whether a prompt will generate an image included in the training data. Prior to presenting this method, we present the definition of image memorization suggested in 
\citep{somepalli2023understanding,carlini2023extracting}. 
\begin{dfn}[$\tau$-Image Memorization]
Given a train set $\mathcal{D}_{train}=\{(\rvx_{train,i}, \rvp_{train,i})\}_{i=1}^N$, a generated image $\rvx$ from a diffusion model $\veps_\vtheta$ trained on $\mathcal{D}_{train}$, and a similarity measurement score SSCD~\citep{pizzi2022self}, image memorization of $\rvx$ is defined as:
    \begin{align}
    \mathcal{M}_\tau(\mathbf{x}, \mathcal{D}_{train}) = \mathbb{I}\left[\exists \, \mathbf{x}_{train} \in \mathcal{D}_{train} \;\text{s.t.}\;\text{SSCD}(\mathbf{x}, \mathbf{x}_{train}) > \tau\right],
    \end{align}
where $\tau$ is a threshold, $\mathbb{I}$ is indicator function, and $\mathcal{M}(\cdot)$ indicates whether the image is memorized.
\end{dfn}
The prior works~\citep{carlini2023extracting,wen2024detecting,webster2023reproducible}  found that prompts inducing memorized images do so regardless of the initial noise, $\rvx_T$, \ie, repeatedly generating the same or almost identical images despite different $\rvx_T$. To quickly identify this case, \citet{wen2024detecting} propose a measure to predict whether a prompt will induce a memorized image using only the first step of the diffusion model, without generating the image. This measure, referred to as $D_\vtheta$, is formulated as follows:
\begin{align}
D_\vtheta(\rvp) = \mathbb{E}_{\rvx_T\sim \mathcal{N}(0,\rmI)}[||\veps_\vtheta(\rvx_T, \rvf(\rvp), T) - \veps_\vtheta(\rvx_T, \rvf(\emptyset), T)||_2].
\end{align}
In this context, the larger $D_{\vtheta}(\rvp)$, the higher the probability that the image generated by the prompt is included in the training data. Denoting image $\rvx$ generated from diffusion model $\veps_\vtheta$ with prompt $\rvp$ as $\rvx(\veps_\vtheta,\rvp)$, we re-purpose it by expressing as 
% this can be expressed as
$D_\vtheta(\rvp)\propto\mathbb{E}[\mathcal{M}(\rvx(\veps_\vtheta,\rvp),\mathcal{D}_{train})]$, where we omit $\tau$ for simplicity. 
The AUC for measuring whether a prompt is a memorized image trigger prompt is reported as 0.960 and 0.990 when the number of initial noises is 1 and 4, respectively.
% , and 0.990 when the number of initial noises is 4. 
% Henceforth, we will refer to this as EDM loss.

\subsection{Memorization Trigger Prompt Searching as an Optimization Problem}\label{sec:optimization}

Our objective is to construct a memorized image trigger prompts dataset and verify corresponding memorized images, \ie to construct $\mathcal{D}_{mem}=\{\rvp\mid\mathbb{E}[\mathcal{M}(\rvx(\veps_\vtheta,\rvp),\mathcal{D}_{train})]>\kappa, \rvp\in\mathcal{T}\}$ where $\kappa$ is the threshold and $\mathcal{T}$ is space of all possible prompts. As mentioned in Section~\ref{related_work}, the prior works~\citep{carlini2023extracting,webster2023reproducible} utilized $\mathcal{D}_{train}$ to search for candidate prompts that could become $\mathcal{D}_{mem}$. They then generated images from these candidate prompts and conducted image retrieval to find memorized images within $\mathcal{D}_{train}$ which is expensive. Moreover, since the training dataset, LAION, is no longer accessible, 
% no longer available,
this approach becomes infeasible.
% However, with the LAION becoming private, this approach is no longer feasible.
% Therefore, 
Thus, we 
% aim to
approach the problem from a different perspective. We search for candidate prompts that could become $\mathcal{D}_{mem}$ without using $\mathcal{D}_{train}$. Then, we generate images from these candidate prompts and use a Reverse Image Search API\footnote{\url{https://tineye.com/}} 
% on the generated images 
to find images on the web akin to generated ones by regarding the web as the training set.
% that are likely from the training data. 
Finally, we perform a human verification process. 

Given that $D_\vtheta(\rvp)\propto\mathbb{E}[\mathcal{M}(\rvx(\veps_\vtheta,\rvp),\mathcal{D}_{train})]$, constructing $\mathcal{D}_{mem}$ can be conceptualized as an optimization problem where we treat the prompt space as a reparametrization space and aim to find prompts yielding high $\mathcal{D}_\vtheta(\rvp)$. 
To formulate the optimization problem, we define the prompt space. Given a finite set $\mathcal{W}$ containing all possible words (tokens), where $|\mathcal{W}| = m$, we model a sentence $\rvp$ with $n$ words as an ordered tuple drawn from the Cartesian product of $\mathcal{W}$, represented as $\mathcal{P} = \mathcal{W}^n$. To solve the optimization problem, we treat $D_\vtheta(\cdot)$ as a negative energy function and model the target Boltzmann distribution $\pi$ such that higher values of $D_\vtheta(\cdot)$ correspond to higher probabilities as
\begin{align}    
\pi(\rvp) = \tfrac{e^{D_\vtheta(\rvp)/K}}{Z},\label{eq:target_distribution}
\end{align}
where $Z = \sum_{\rvp \in \mathcal{P}}{e^{D_\vtheta(\rvp)/K}}$ is a regularizer and $K$ is a temperature constant. By sampling from modeled target distribution $\pi(\rvp)$ in a discrete, finite, multivariate, and non-differentiable space $\mathcal{P}$, we can obtain prompts that maximize $D_\vtheta(\rvp)$, which are likely to be memorized image trigger prompts.

\subsection{Constructing MCMC Leveraging $D_\theta$}
To tackle the aforementioned challenging optimization problem, we propose to use  Markov Chain Monte Carlo (MCMC)~\citep{hastings1970monte} to sample from the target distribution $\pi(\rvp)$. This method allows us to efficiently explore the discrete prompt space and find prompts likely to induce memorized images, effectively navigating $\mathcal{P}$ to identify optimal prompts. From any arbitrary distribution of sentence, $\pi_0$, Markov Chain with transition matrix $\rmT$ can be developed as follows:
\begin{align}
\pi_{i+1}=\pi_i\rmT.
\end{align}
It is well known that Markov Chains satisfying irreducibility and aperiodicity converge to certain distribution $\pi^*$~\citep{robert1999monte}, which can be formulated as $\pi_n=\pi_0\rmT^n\rightarrow\pi^*$ independent of $\pi_0$. In this context, the transition matrix can vary depending on the algorithm used to solve the MCMC. By carefully choosing the sampling algorithm, we can ensure that the final distribution $\pi^*$ reached by the transition matrix converges to desired target distribution $\pi$~\citep{robert1999monte,geman1984stochastic,hastings1970monte}. 
%The most well-known algorithm is Metropolis-Hastings~\citep{hastings1970monte} algorithm, which has been proven to have its transition matrix converge to the target distribution~\citep{robert1999monte}. However, due to the multi-dimensional nature of our parameter space, 
%the Metropolis-Hastings algorithm converges slowly. Therefore, 
% Considering the multi-dimensional nature of our parameter space, we employ the Gibbs sampling algorithm~\citep{geman1984stochastic}. Gibbs sampling algorithm also has proven convergence of the transition matrix and is known for faster convergence in multi-dimensional problems~\citep{johnson2013component,liu2000generalised}.
Considering the multi-dimensional nature of our parameter space, we employ the Gibbs sampling algorithm~\citep{geman1984stochastic} for simplicity.
% \footnote{One can deploy and explore more advanced sampling techniques, but we leave that for future work.}
Gibbs sampling is an MCMC sampling algorithm method where, at each step, only one coordinate of the multi-dimensional variable is updated to transition from the current state to the next state. Gibbs sampling algorithm has proven the convergence of the transition matrix and is known for fast convergence in multi-dimensional problems~\citep{johnson2013component,terenin2020asynchronous,papaspiliopoulos2008stability}.
We adopt random scan Gibbs sampling, which involves randomly selecting an index and updating the value at that index. This process can be expressed as the sum of $n$ transition matrices, as follows:
\begin{align}
    \rmT &= \sum_{i=1}^n{\frac{1}{n} \cdot \rmT_i}, \\
    [\rmT_i]_{\rvp^j \rightarrow \rvp^{j+1}} &= 
    \left\{
    \begin{array}{cl}
        \pi(\rvp^{j+1}_i|\rvp^{j}_{-i}) & \text{if } \rvp_{-i}^j = \rvp^{j+1}_{-i} \\
        0 & \text{else},
    \end{array}
    \right.
\end{align}
where $\rvp_{-i}=\{\rvp_1,\rvp_2,...,\rvp_{i-1},\rvp_{i+1},...,\rvp_n\}$ and $\rvp^j$ is a $j$-th state prompt. Integrating Equation~\ref{eq:target_distribution} into the above formulas, the final transition matrix is obtained as follows:
\begin{align}
    [\rmT]_{\rvp^j\rightarrow \rvp^{j+1}} = \left\{\begin{array}{ll}
      \frac{1}{n}\cdot( \frac{e^{D_\vtheta(\mathcal{P}_i =\rvp_i^{j+1},\mathcal{P}_{-i}=\rvp_{-i}^j)/K}}{\sum_{\mathbf{w} \in \mathcal{W}} e^{D_\vtheta(\mathcal{P}_i =\mathbf{w},\mathcal{P}_{-i}=\rvp_{-i}^j)/K}}) & if\;\rvp_{-i}^j=\rvp_{-i}^{j+1},\label{eq:final_transition}\\
      0 & else,\\
\end{array} 
\right.%\\
% \pi_n=\pi_0P^n\rightarrow\pi
\end{align}
where detailed derivation is provided in Appendix. Since it is impractical to compute $D_\vtheta(\cdot)$ for all $\rvw \in \mathcal{W}$, we approximate $\mathcal{W}$ as top $Q$ samples obtained from BERT~\citep{devlin2018bert}. This means that the $i$-th element of the prompt $\rvp$ is masked and BERT is used to predict the word, from which the top Q samples are selected as candidate words. Mathematical derivation is complex, but the algorithm is straightforward: the process iteratively 1) selects and replace a word into [MASK] token from the sentence, 2) predicts top $Q$ words via BERT and computes proposal distribution, and 3) stochastically replaces it according to the proposal distribution. Please refer to Algorithm \ref{algorithm} for details.

\begin{algorithm}[t]
\caption{Memorized Image Trigger Prompt Searching via Gibbs sampling}
\label{algorithm}
\begin{algorithmic}[1]
\State \textbf{Input:} Diffusion model $\vtheta$, BERT model $\vphi$, initial sentence $\rvp^0$ with length $n$, iteration number $N$, number of proposal words $Q$, termination threshold $\kappa$, hyperparameter $K$, $\gamma$, $\{r_1, \ldots, r_n\}$
\State $\rvp^* \gets \rvp^0$
\While{$D_\vtheta(\rvp^*) < \kappa$}
    \For{$j = 0$ to $N$}
        \State Randomly select index $i \in \{1, \ldots, n\}$
        \State $\mathcal{W}_Q \gets \operatorname*{arg\,top}_Q \; p_\vphi(\rvw \mid \rvp_{-i}^j)$
        \State $p(\rvp_i^{j+1} \mid \rvp_{-i}^j) \gets \frac{e^{D_\vtheta(\mathcal{P}_i =\rvp_i^{j+1},\mathcal{P}_{-i}=\rvp_{-i}^j)/K}}{\sum_{\mathbf{w} \in \mathcal{W}_Q} e^{D_\vtheta(\mathcal{P}_i =\mathbf{w},\mathcal{P}_{-i}=\rvp_{-i}^j)/K}}$
        \State $\rvp_{i}^{j+1} \gets \text{Sample from } p(\rvp_i^{j+1} \mid \rvp_{-i}^j)$
        \State $\rvp^{j+1} \gets (\rvp_1^j, \rvp_2^j, \ldots, \rvp_{i}^{j+1}, \ldots, \rvp_n^j)$
    \EndFor
    \State $\rvp^* \gets \argmax_{\rvp\in\{\rvp^0,\rvp^1,...,\rvp^n\}}D_\vtheta(\rvp)$
\EndWhile
\State \textbf{return} $\rvp^*$
\end{algorithmic}
\end{algorithm}

\subsection{Dataset Construction Leveraing MCMC}\label{construction}
We conduct dataset construction in two stages: 1) using a masked sentence as the prior and employing MCMC to find memorized image trigger prompts, and 2) using the memorized image trigger prompts as the prior for augmentation through MCMC.

\paragraph{Using Masked Sentence as Prior.}
This stage aims to discover new memorized images. The sentence is initialized with sentence of length $n$ [MASK] token, \ie $\rvp_0 =$ \{[MASK], [MASK], ... , [MASK]\}. We then employ Algorithm~\ref{algorithm} initialized with $\rvp_0$ to obtain the candidate prompt. Similar to the conventional approach~\citep{carlini2023extracting} to extract train images, we then generate 100 images for this prompt and leverage DBSCAN~\citep{ester1996density} clustering algorithm with SSCD~\citep{pizzi2022self} to extract images forming at least 20 nodes. Those images are employed to Reverse Image Search API to find train image sources and human verification is conducted.

\paragraph{Using Found Trigger Prompts as Prior.}
This stage aims to augment memorized image trigger prompts. We leverage the prompts found in the previous stage or those provided by \citet{webster2023reproducible} as the prior, $\pi_0$. 
In this process, we employ a slightly modified algorithm to enhance diversity. Instead of running a single chain for one prompt, we run $n$ separate chains for each word position in an $n$-length sentence, treating each position as the first updating index in Gibbs sampling. This method ensures a varied exploration of the prompt space. We then save the top 100 prompts with the highest $D_\vtheta(\cdot)$. 
%For these prompts, we generate 10 images each and filter the prompts that generate images with the highest SSCD to the dataset. 
We retained all prompts generated during the MCMC sampling process and then selected 20 augmented prompts per original prompt, considering diversity.
The detailed process is provided in Appendix.

%% file: 4_result.tex
\section{Dataset Analysis and Efficiency of Trigger Prompt Searching Algorithm}\label{dataset}

\subsection{Dataset Analysis}

\begin{table}[t]
\centering
\scalebox{0.88}{
\begin{tabular}{lcccccccc}
\toprule
& \multicolumn{2}{c}{Stable Diffusion 1} & \multicolumn{2}{c}{Stable Diffusion 2} & \multicolumn{2}{c}{DeepFloydIF} & \multicolumn{2}{c}{Realistic Vision} \\
\cmidrule(lr){2-3} \cmidrule(lr){4-5} \cmidrule(lr){6-7} \cmidrule(lr){8-9}
& \makecell{Trigger \\ Prompt \#} & \makecell{Mem. \\ Image \#} & \makecell{Trigger \\ Prompt \#} & \makecell{Mem. \\ Image \#} & \makecell{Trigger \\ Prompt \#} & \makecell{Mem. \\ Image \#} & \makecell{Trigger \\ Prompt \#} & \makecell{Mem. \\ Image \#} \\
\midrule
\citet{webster2023reproducible} & 325 & 111 & 210 & 25 & 162 & 17 & 354 & 119 \\
\textbf{MemBench} & \textbf{3000} & \textbf{151} & \textbf{1500} & \textbf{55} & \textbf{309} & \textbf{51} & \textbf{1352} & \textbf{148} \\
\bottomrule
\end{tabular}
}
\vspace{10pt} 
\caption{Comparison of the number of memorized images and trigger prompts provided in each dataset. Our dataset is significantly larger in terms of the number of trigger prompts across all models. Please note that images sharing the same layout, as depicted in Figure~\ref{fig:layout_duplication}, have been counted as a single image.}
\label{tb:comparison}
\end{table}

\clearpage

\begin{wrapfigure}{r}{0.27\linewidth}
   \includegraphics[width=\linewidth]{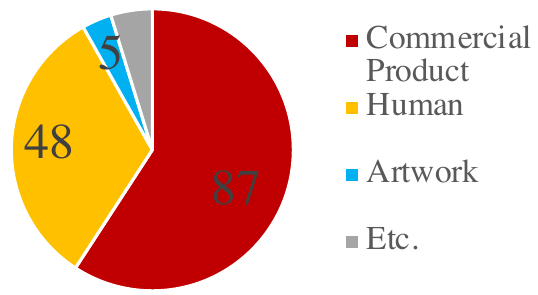}
   \vspace{-0.5cm}
   \caption{Components of Memorized Images in Stable Diffusion 1}
\label{fig:data_component}
\vspace{-0.2cm}
\end{wrapfigure}

Table~\ref{tb:comparison} presents the number of memorized images and trigger prompts obtained using the methodology described in Section~\ref{method}. For both Stable Diffusion 1 and 2, the number of prompts has increased more than fivefold for each model and more than 9 times in total compared to those reported by \citet{webster2023reproducible}. The number of memorized images included in the dataset has also increased, with Stable Diffusion 2 showing an increase of over twofold. Additionally, we provide memorized images and trigger prompts for DeepFloydIF~\citep{shonenkov2023deepif}, which has a cascaded structure, and Realistic Vision~\citep{CivitAI}, an open-source diffusion model. For these two models as well, we provide a greater number of memorized images and trigger prompts than Webster et al. We have also applied our algorithm to the more recent model, Stable Diffusion 3~\citep{esser2024sd3}. Please refer to Appendix for the results. The composition of the images included in MemBench is shown in Figure~\ref{fig:data_component}, illustrating that the memorized images encompass a substantial number of commercial product images and human images. It also includes artwork such as brand logos.

\subsection{Efficacy of Memorized Image Trigger Prompt Searching}\label{result:algorithm}
\begin{table}[t]
\centering
\begin{tabular}{@{}lccccc@{}}
\toprule
 & Greedy Search & ZeroCap & PEZ & ConZIC & \textbf{Ours} \\ \midrule
Hours/Memorized Image & 5.7 & None & None & 3.81 & \textbf{2.08} \\ \bottomrule
\end{tabular}
\vspace{10pt}
\caption{Comparison of the efficiency of our method and other prompt space optimization methods. Experiment was done on 1 A100 GPU}
\label{searching_efficacy}
\end{table}

In this section, we validate the efficiency of our method in discovering memorized images without access to $\mathcal{D}_{train}$. The task of finding memorized image trigger prompts without $\mathcal{D}_{train}$ is defined as follows: without any prior information, the method must automatically find trigger prompts that induce memorized images. This involves: 1) selecting candidate prompts, 2) generating 100 images for each candidate prompt, 3) applying DBSCAN~\citep{ester1996density} clustering with SSCD to get candidate images and using a Reverse Image Search API to verify those images' presence on the web. To the best of our knowledge, this task is novel, so we provide naive baselines. As the first baseline, we perform a greedy search by measuring $D_\theta$ for all prompts in the prompt dataset and selecting the top 200 prompts as candidate prompts. For the prompt dataset, we leveraged DiffusionDB, which contains 13M prompts collected from diffusion model users. Additionally, we provide three other baselines, all of which are algorithms that solve optimization problems in the prompt space. For two of these baselines, we adapt ZeroCap~\citep{tewel2022zerocap} and ConZIC~\citep{zeng2023conzic}, methods designed to maximize the CLIP Score for zero-shot image captioning, by replacing their objective function with $D_\theta$. Similarly, we also adapt PEZ~\citep{wen2024hard} by substituting its objective function with $D_\theta$ to serve as another baseline. For each of these three methods, we conducted 200 iterations and obtained prompts. For our method, we performed 200 MCMC runs with 150 iterations each, and selected the resulting prompts as candidate prompts. For more detailed implementation, please refer to Appendix.

The results are shown in Table~\ref{searching_efficacy}. Our method significantly outperforms other methods. Comparing with baselines~\citep{carlini2023extracting,webster2023reproducible} that leverage LAION itself is challenging, as the dataset is no longer available and the elements for reimplementation are omitted in the corresponding papers. However, as mentioned in Section~\ref{related_work}, their memory-inefficient and computationally intensive methods provided only few memorized images and trigger prompts. Please refer to Appendix for the performance of augmenting the memorized image trigger prompts through MCMC.

%% file: 5_membench.tex
\section{MemBench: Metrics, Scenarios and Reference Performance}\label{membench}
\begin{figure}[t]
\begin{center}
   \includegraphics[width=\linewidth]{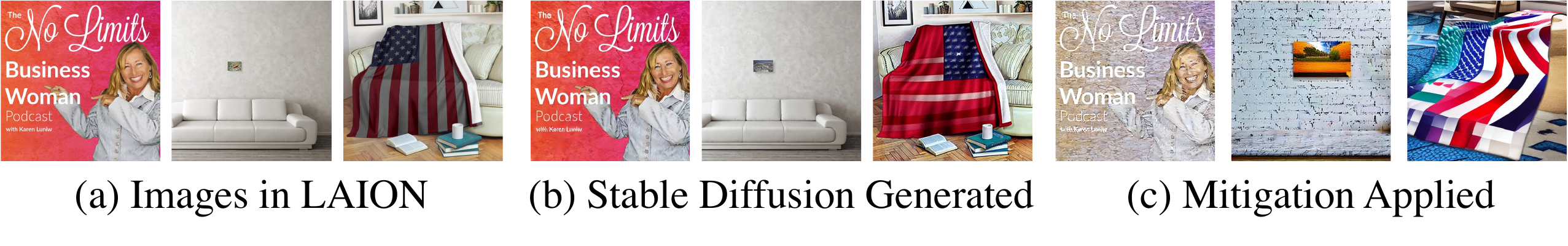}
\end{center}
\vspace{-0.5cm}
    \caption{The necessity of measuring the Aesthetic score. Images generated with the mitigation method applied are not desirable but achieve a low SSCD while maintaining a high CLIP Score.}
\label{fig:image_quality}
\end{figure}

\paragraph{Metrics.}\label{metrics}
We present rigorous metrics for correctly evaluating mitigation methods, which include similarity score, Text-Image alignment score, and quality score. Firstly, following previous works, we adopt \textbf{SSCD}~\citep{pizzi2022self} as the similarity score and measure max SSCD between a generated image using trigger prompt and memorized images. In detail, if a prompt $\rvp^*$ triggers images $\{\rvx_{1}^*, ..., \rvx_{k}^*\}$ included in $\mathcal{D}_{train}$, we measure $\max_{\rvx \in \{\rvx_{1}^*, ..., \rvx_{k}^*\}} SSCD(\rvx(\rvp^*,\veps_\theta), \rvx)$. Secondly, we adopt \textbf{CLIP Score}~\citep{hessel2021clipscore} to measure Text-Image alignment between prompt and generated images. Lastly, We adopt an \textbf{Aesthetic Score}~\citep{schuhmann2022laion} as the image quality score. While previous works did not measure image quality scores, we observed issues shown in Figure~\ref{fig:image_quality}. When memorization mitigation methods are applied, we observed that 
% it has been observed that 
image quality degrades, the rich context generated by the diffusion model is destroyed, or distorted images are formed. An ideal memorization mitigation method should be able to preserve the image generation capabilities of the diffusion model.

\paragraph{Scenarios.} To ensure that memorization mitigation methods can be generally applied to diffusion models, we provide two scenarios: the memorized image trigger prompt scenario and the general prompt scenario. First, the memorized image trigger prompt scenario evaluates whether mitigation methods can effectively prevent the generation of memorized images. This scenario uses the memorized image trigger prompts we identified in Section~\ref{method}. In this scenario, we generate 10 images for each trigger prompt and measure the Top-1 SSCD and the mean values of the Top-3 SSCD. We also measure the proportion of images with SSCD exceeding 0.5. For CLIP Score and Aesthetic Score, we calculate the average value across all generated images. Second, the general prompt scenario ensures that the performance of the diffusion model does not degrade when using prompts other than trigger prompts. We leverage the COCO~\citep{lin2014microsoft} validation set as general prompts. In this scenario, images are generated once per prompt, and the average CLIP Score and Aesthetic Score are measured.

\paragraph{Reference Performance.} We propose a reference performance for interpreting the performance of mitigation methods. 
An effective mitigation method should be able to reduce SSCD while maintaining CLIP Score. However, although SSCD is a metric designed to compare the structural similarity of images for copy detection tasks, it inevitably includes semantic meaning due to the self-supervised nature of the trained neural network. On the other hand, the semantic meaning of the trigger prompt should still be reflected in the generated image to maintain CLIP Score even when a mitigation method is applied. 
Therefore, it is uncertain how much the SSCD between memorized images and generated images can be reduced while maintaining the CLIP Score between trigger prompts and generated images.
In this regard, we provide a reference performance to indicate how much SSCD can be reduced while maintaining a high CLIP Score. We assume querying images with trigger prompts via the Google Image API\footnote{\url{https://developers.google.com/custom-search}} as a strong proxy model for the generative model and provide the reference performance based on this approach. Please refer to Appendix for details.

%% file: 6_findings.tex
\section{Deeper Analysis into Image Memorization}

\begin{figure}[t]
    \centering
    \begin{minipage}[t]{0.2\textwidth}
        \centering
        \includegraphics[height=0.1\textheight, keepaspectratio]{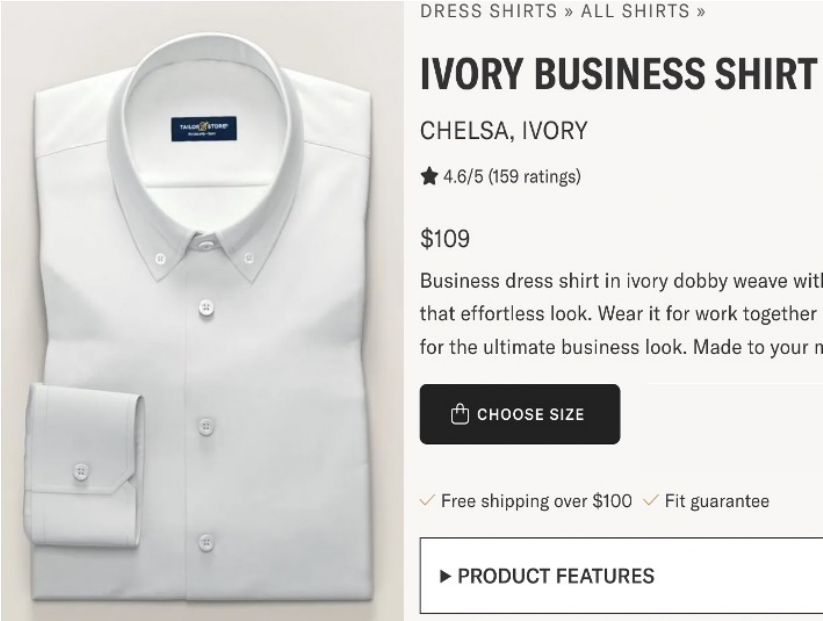}
        \caption*{(a)}
    \end{minipage}
    \hfill
    \begin{minipage}[t]{0.16\textwidth}
        \centering
        \includegraphics[height=0.1\textheight, keepaspectratio]{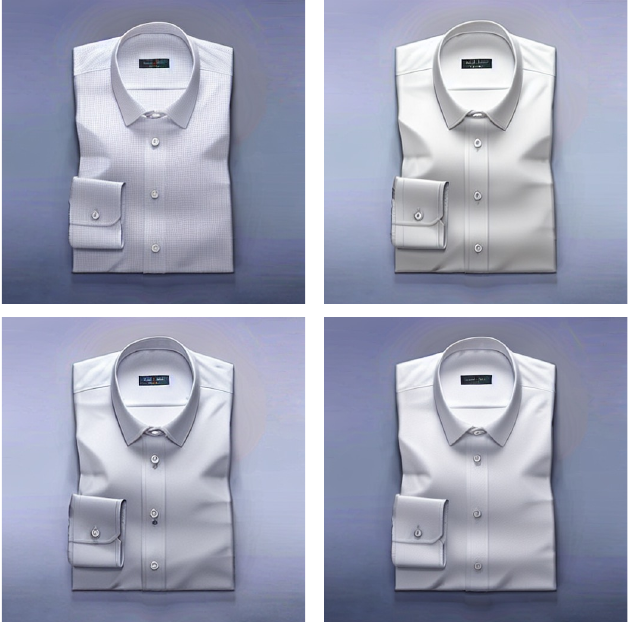}
        \caption*{(b)}
    \end{minipage}
    \hfill
    \begin{minipage}[t]{0.32\textwidth}
        \centering
        \includegraphics[height=0.1\textheight, keepaspectratio]{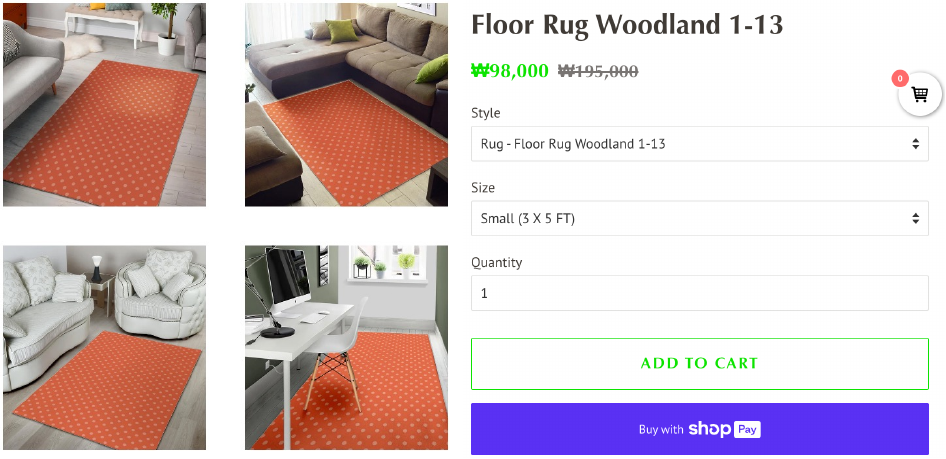}
        \caption*{(c)}
    \end{minipage}
    \hfill
    \begin{minipage}[t]{0.16\textwidth}
        \centering
        \includegraphics[height=0.1\textheight, keepaspectratio]{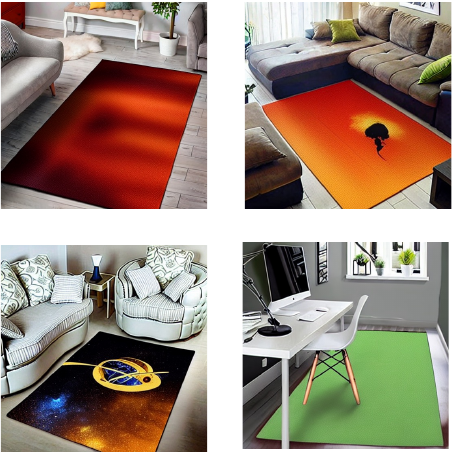}
        \caption*{(d)}
    \end{minipage}
    \caption{Examples of memorized images found  using the Reverse Image Search API. (a), (c) Shirt/rug currently sold commercially, (b), (d) four images generated by Stable Diffusioon}
    \label{fig:examples}
\end{figure}

\begin{figure}[t]
\begin{center}
   \includegraphics[width=0.9\linewidth]{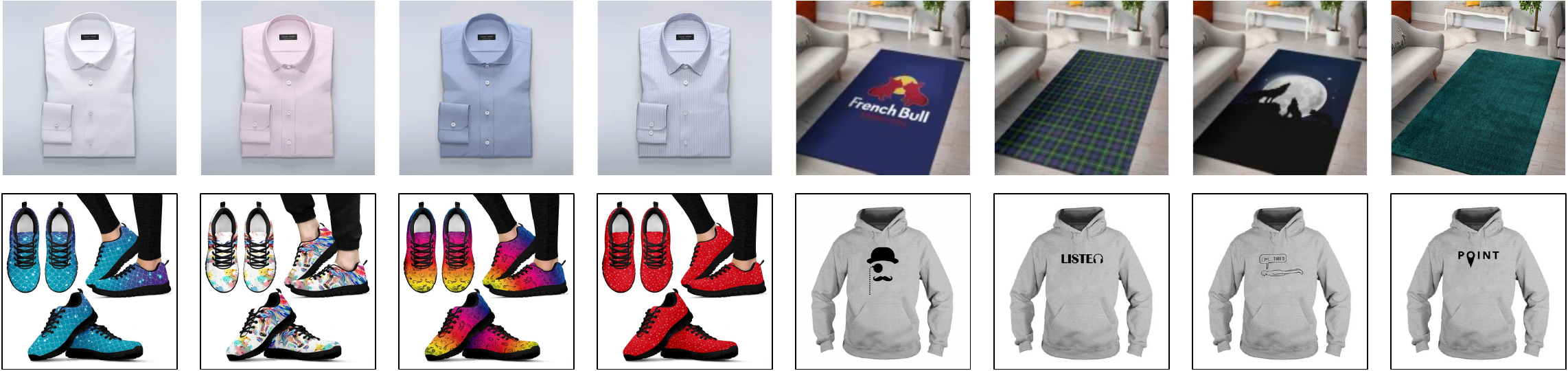}
\end{center}
    \caption{Results of images found by leveraging Reverse Image Search API to the images generated from trigger prompts. The shared layout suggests the occurrence of image memorization.}
\label{fig:layout_duplication}
\end{figure}

In this section, we provide a deeper analysis of the memorized images and trigger prompts in MemBench. Firstly, we have found that \textbf{Stable Diffusion regenerates commercial products currently on sale}. While the possibility that diffusion models could memorize commercial images has been suggested~\citep{carlini2023extracting,somepalli2023diffusion}, we are the first to confirm this. Unlike the previous studies~\citep{carlini2023extracting,webster2023reproducible} that used image retrieval from LAION to find memorized images, we leverage a Reverse Image Search API to find those, which enable us this verification. As shown in Figure~\ref{fig:examples}.b, Stable Diffusion replicates images of commercially available shirts when given a specific prompt. Figure~\ref{fig:examples}.d further illustrates the replication of layouts; for a commercially sold carpet, all layouts have been reproduced.

Secondly, we explore the cause of image memorization in Stable Diffusion 2, trained on LAION-5B, whose duplicates are removed. Previous works~\citep{somepalli2023understanding, gu2023memorization} suggested that image memorization issues arise from duplicate images in the training data. \citet{webster2023duplication} confirmed that the LAION-2B dataset contains many duplicate images likely to be memorized. However, Stable Diffusion 2 still exhibits image memorization issues while reduced. We hypothesize that this memorization arises due to layout duplication. Figure~\ref{fig:layout_duplication} shows the images found by Reverse Image Search API that are memorized by Stable Diffusion. We found that there are often over 100 images on the web with the same layout but different color structures. LAION-5B underwent deduplication based on URLs\footnote{\url{https://laion.ai/blog/laion-5b/}}, but this process may not have removed these images. 
These layout memorizations are also obviously subject to copyright, posing potential social issues. Additional examples are provided in Appendix.

%% file: 7_evaluation.tex
\section{Evaluation of Image Memorization Mitigation Method on MemBench}\label{evaluation}

\begin{table}[t]
    \centering
    \captionsetup{font=normalsize} 
    \scalebox{0.68}{
    \begin{tabular}{llccccccc}
        \toprule
        & & \multicolumn{5}{c}{MemBench} & \multicolumn{2}{c}{COCO} \\
        \cmidrule(lr){3-7} \cmidrule(lr){8-9}
        & & Top-1 SSCD $\downarrow$& Top-3 SSCD $\downarrow$& SSCD > 0.5 $\downarrow$& CLIP $\uparrow$& Aesthetic $\uparrow$& CLIP $\uparrow$& Aesthetic $\uparrow$\\
        \midrule
        Base &  & 0.641 & 0.605 & 0.451 & 0.273 & 5.25 & 0.321 & 5.37 \\
        \textbf{Reference Performance}  &\hspace{-0.2cm}(API search)  & \textbf{0.088} & - & - & \textbf{0.310} & - & - & - \\
        \midrule
        \multirow{5}{*}[1.5ex]{\shortstack[l]{RNA\citep{somepalli2023understanding}}} & n = 1 & 0.479 & 0.425 & 0.241 & 0.270 & 5.18 & 0.314 & 5.34 \\
                             & n = 2 & 0.389 & 0.338 & 0.165 & 0.270 & 5.14 & 0.310 & 5.33 \\
                             & n = 3 & 0.329 & 0.280 & 0.121 & 0.267 & 5.13 & 0.307 & 5.30 \\
                             & n = 4 & 0.287 & 0.239 & 0.089 & 0.264 & 5.10 & 0.304 & 5.29 \\
                             & n = 5 & 0.254 & 0.213 & 0.074 & 0.262 & 5.08 & 0.302 & 5.28 \\
                             & n = 6 & 0.228 & 0.189 & 0.055 & 0.258 & 5.06 & 0.298 & 5.24 \\
        \midrule
        \multirow{5}{*}[1.5ex]{\shortstack[l]{RTA \citep{somepalli2023understanding}}} & n = 1 & 0.497 & 0.446 & 0.265 & 0.269 & 5.20 & 0.316 & 5.34 \\
                             & n = 2 & 0.397 & 0.347 & 0.175 & 0.268 & 5.19 & 0.314 & 5.32 \\
                             & n = 3 & 0.330 & 0.285 & 0.129 & 0.266 & 5.17 & 0.310 & 5.29 \\
                             & n = 4 & 0.282 & 0.240 & 0.094 & 0.264 & 5.15 & 0.306 & 5.27 \\
                             & n = 5 & 0.257 & 0.217 & 0.080 & 0.262 & 5.14 & 0.302 & 5.26 \\
                             & n = 6 & 0.228 & 0.190 & 0.056 & 0.258 & 5.10 & 0.299 & 5.27 \\
        \midrule
        \multirow{5}{*}[1.5ex]{\citet{wen2024detecting}} & l = 7 & 0.410 & 0.346 & 0.134 & 0.270 & 5.16 & 0.321 & 5.37 \\
                                     & l = 6 & 0.355 & 0.289 & 0.089 & 0.270 & 5.15 & 0.321 & 5.37 \\
                                     & l = 5 & 0.312 & 0.246 & 0.059 & 0.269 & 5.14 & 0.321 & 5.37 \\
                                     & l = 4 & 0.259 & 0.199 & 0.035 & 0.268 & 5.13 & 0.321 & 5.37 \\
                                     & l = 3 & 0.181 & 0.139 & 0.015 & 0.264 & 5.11 & 0.321 & 5.37 \\
                                     & l = 2 & 0.096 & 0.075 & 0.001 & 0.242 & 4.97 & 0.321 & 5.37 \\
        \midrule
        \multirow{4}{*}[1.5ex]{\citet{ren2024unveiling}} & c = 1.0 & 0.289 & 0.247 & 0.083 & 0.263 & 5.17 & 0.316 & 5.33 \\
                                     & c = 1.1 & 0.283 & 0.239 & 0.071 & 0.260 & 5.17 & 0.313 & 5.31 \\
                                     & c = 1.2 & 0.278 & 0.232 & 0.058 & 0.257 & 5.15 & 0.309 & 5.28 \\
                                     & c = 1.3 & 0.275 & 0.227 & 0.050 & 0.254 & 5.14 & 0.304 & 5.26 \\
        \bottomrule
    \end{tabular}
    }
    \vspace{0.2cm}
    \caption{Performance evaluation of image memorization mitigation methods in MemBench. 
    Please refer to Appendix for the details of hyper-paramters.}
    \label{tb:evaluation}
    \vspace{-0.5cm}
\end{table}

In this section, we evaluate image memorization mitigation methods on our MemBench in Stable Diffusion 1. For results of Stable Diffusion 2, please refer to Appendix.

\paragraph{Baselines.} We use Stable Diffusion 1.4 as the base model. The image memorization mitigation methods evaluated include: 1) RTA~\citep{somepalli2023understanding}, which applies random token insertion to the prompt, 2) RNA~\citep{somepalli2023understanding}, which inserts a random number between $[0, 10^6]$ into the prompt, 3) method proposed by \citet{wen2024detecting} that applies adversarial attacks to text embeddings, and 4) method proposed by \citet{ren2024unveiling} that rescales cross-attention. Image generation is performed using the DDIM~\citep{song2022denoising} Scheduler with a guidance scale of 7.5 and 50 inference steps.

\paragraph{Results.} We present the experimental results in Table~\ref{tb:evaluation}. As shown in Table~\ref{tb:evaluation}, for all methods, lowering the SSCD significantly reduces both the CLIP Score and the Aesthetic Score. This indicates a degradation in text-image alignment and image quality. In particular, upon examining images with low Aesthetic Scores, we observe that issues in Figure~\ref{fig:image_quality} occur across all methods. While \citet{ren2024unveiling} measured FID, the Aesthetic Score offers a more straightforward way to evaluate individual image quality and better highlight these issues. As reported by \citet{wen2024detecting}, all methods exhibit a trade-off between SSCD and CLIP Score. Regarding the reference performance obtained via API search, it can be observed that the SSCD can be reduced to 0.088 while maintaining a high CLIP Score. Due to the inherent limitations of the Stable Diffusion baseline model, the CLIP Score cannot exceed 0.273 when mitigation methods are applied. However, mitigation methods should aim to reduce the Top-1 SSCD to around 0.088 while maintaining at least this level of CLIP Score.

To provide a more detailed analysis of each method, we observe that the approach proposed by Wen~et al. achieves the best performance in the trade-off between SSCD and CLIP Score. However, to reduce the proportion of images with SSCD exceeding 0.5—indicative of image memorization—to nearly zero, their method still requires a reduction in CLIP Score by 0.025. Given the scale of the CLIP Score, this drop suggests that the generated images may be only marginally related to the given prompts. Moreover, a significant decrease in the Aesthetic Score is also observed. On the other hand, the method proposed by Wen et al. has an additional advantage: it does not result in any performance drop in the general prompt scenario on the COCO dataset, making it the most suitable option for practical applications as of now.

The most recent method proposed by \citet{ren2024unveiling} shows a considerable reduction in the CLIP Score. Even at the lowest hyper-parameter setting ($c=1.0$), the reductions in both CLIP Score and Aesthetic Score are substantial, limiting its general applicability to diffusion models. 
The most basic approaches, RNA and RTA, show a decrease in CLIP Score by 0.015 at the hyper-parameter setting ($n=6$) that lowers the proportion of images with SSCD exceeding 0.5 to 0.05. This result is expected given the nature of these methods: both attempt to prevent image memorization by adding irrelevant tokens to the prompts. As a result, RNA and RTA are unreliable for application to diffusion models.

%% file: 8_conclusion.tex
\section{Conclusion}\label{conclusion}
We have presented MemBench, the first benchmark for evaluating image memorization mitigation methods in diffusion models. MemBench includes various memorized image trigger prompts, appropriate metrics, and a practical scenario to ensure that mitigation methods can be effectively applied in practice. We have provided the reference performance that mitigation methods should aim to achieve. Through MemBench, we have confirmed that existing image memorization mitigation methods are still insufficient for application to diffusion models in practical scenarios. The lack of a benchmark may have previously hindered the research of effective mitigation methods. However, we believe that our benchmark will facilitate significant advancements in this field. 

\paragraph{Limitations and Future Work.} Another contribution of our work is providing an algorithm for efficiently searching memorized image trigger prompts based on MCMC. Our approach is faster than other searching algorithms we have tried, yet it does not exhibit exceptionally high speed. Consequently, due to time constraints, we were unable to provide a larger number of memorized images. However, our method allows for the continuous search of more memorized images and their corresponding trigger prompts, and we plan to update the dataset regularly. Additionally, we aim to enhance the efficiency of our memorized image trigger prompt searching algorithm in the future.